\documentclass[conference]{IEEEtran}
\usepackage{times}
\usepackage[numbers,sort&compress]{natbib}
\usepackage{multicol}
\usepackage[bookmarks=true]{hyperref}
\IEEEoverridecommandlockouts
\usepackage{graphicx}
\usepackage{subfig} 
\usepackage[skip=0pt,font=small,labelfont=bf]{caption} 
\usepackage{textcomp}
\usepackage{color}
\usepackage{mathtools}
\usepackage{amsmath}
\usepackage{epsfig,graphicx,amsmath,amsfonts}
\usepackage{xcolor,import}
\usepackage{transparent}
\usepackage{upgreek}

\newcommand\numberthis{\addtocounter{equation}{1}\tag{\theequation}}
\usepackage[ruled,linesnumbered]{algorithm2e}
\usepackage[normalem]{ulem}
\usepackage{multirow}
 
\usepackage{float}
\usepackage[utf8]{inputenc}
\usepackage[TS1,T1]{fontenc}
\usepackage{array, booktabs}
\usepackage{graphicx}
\usepackage{comment}
\usepackage{colortbl}
\usepackage{caption}
\usepackage{amsthm}

\newcommand{\shrinka}{\def\baselinestretch{1.}\large\normalsize}

\title{\Large \bf Geometric In-Hand Regrasp Planning: \\ Alternating Optimization of Finger Gaits and In-Grasp Manipulation}
\author{\authorblockN{Balakumar Sundaralingam and Tucker Hermans}\authorblockA{University of Utah Robotics Center and the School of Computing\\University of Utah, Salt Lake City, UT, USA\\Email: {\tt\small\{bala, thermans\}@cs.utah.edu}}}
\begin{document}
\shrinka
\maketitle
\begin{abstract}
  This paper explores the problem of autonomous, in-hand regrasping--the problem of moving from an initial grasp on an object to a desired grasp using the dexterity of a robot's fingers. We propose a planner for this problem which alternates between finger gaiting, and in-grasp manipulation. Finger gaiting enables the robot to move a single finger to a new contact location on the object, while the remaining fingers stably hold the object. In-grasp manipulation moves the object to a new pose relative to the robot's palm, while maintaining the contact locations between the hand and object. Given the object's geometry (as a mesh), the hand's kinematic structure, and the initial and desired grasps, we plan a sequence of finger gaits and object reposing actions to reach the desired grasp without dropping the object. We propose an optimization based approach and report in-hand regrasping plans for 5 objects over 5 in-hand regrasp goals each. The plans generated by our planner are collision free and guarantee kinematic feasibility.
\end{abstract}
\section{Motivation}
In-hand regrasping, the problem of moving from an initial grasp to a desired grasp on an object without using the environment for support, remains a challenging task for robots. The task involves breaking contacts and making new contacts with the object of interest, while not dropping the object. Humans regrasp objects in-hand everyday, when using precision tools such as a screwdriver, picking up a pen to write, or manipulating a mobile phone to send a text message. Many other tools, such as hammers and screwdrivers, require a specific grasp for operating, but must be grasped differently when picked up in order to avoid collision from the environment. For example, a mobile phone lying flat on a table must be picked up from the edges using the fingertips of the hand and then reoriented into the user's palm, in order to type on the screen.

Endowing robots with the skill of in-hand regrasping would enable them to use many tools common in human environments. Additionally, uncertainty stemming from sensor measurements or lack of knowledge of a specific object, may cause a robot to initially grasp an object differently than intended. Cluttered and constrained environments further restrict the set of feasible grasps, such as those available when grasping an object in a messy drawer. After picking up and removing an object with an available grasp, a robot can move the object into free space and then use in-hand regrasping to switch to a task-specific grasp.
\begin{figure}
  \centering
  \includegraphics[width=0.45\textwidth]{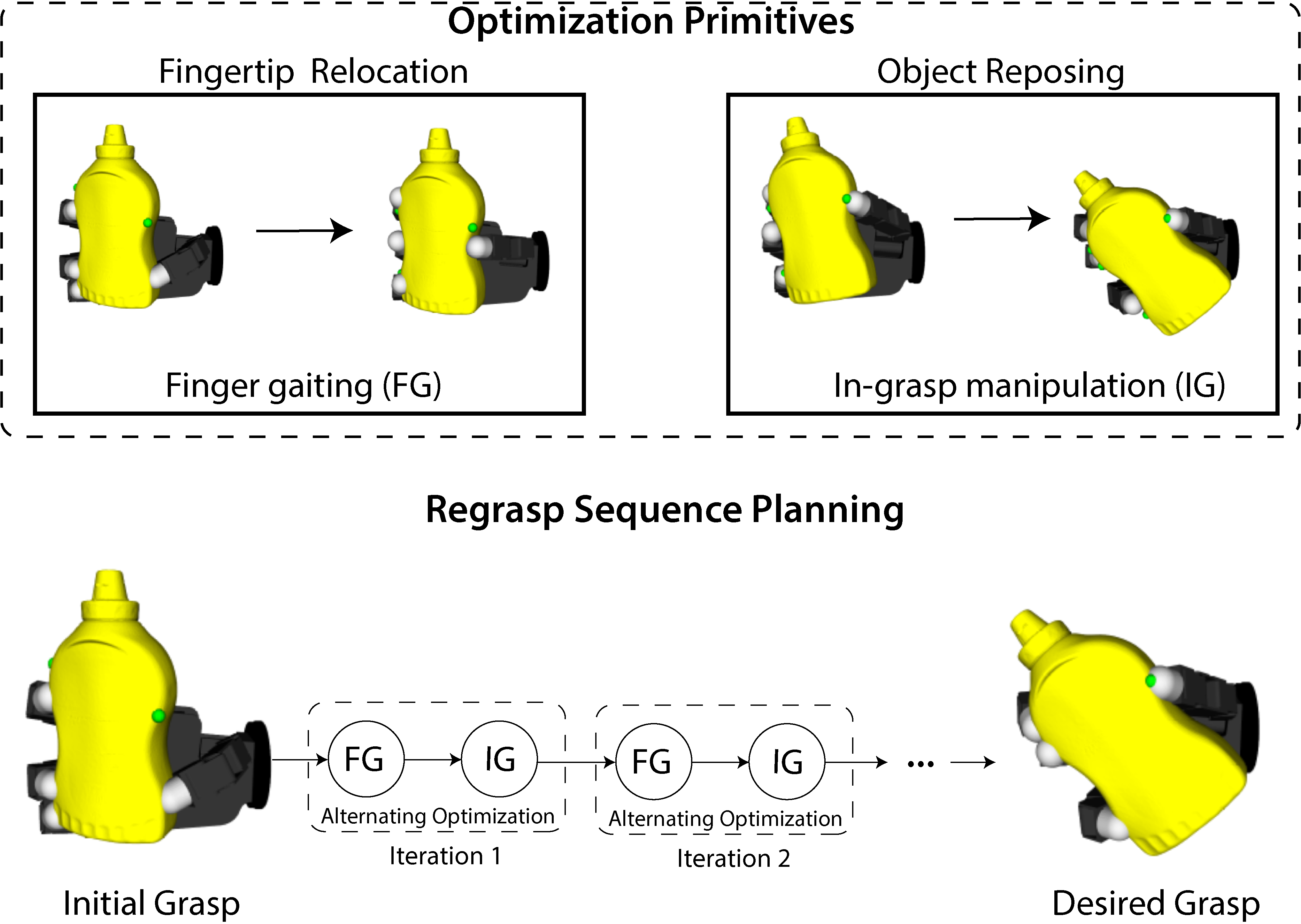}
  \caption{Our planner builds on two in-hand manipulation primitives--finger gaiting and in-grasp manipulation--both formulated as optimization problems. Our regrasping planner must then find an alternating sequence of desired finger contact locations and object poses to move from an initial grasp to the desired grasp using the two primitives.}
  \label{fig:intro_plan}
\end{figure}

In this paper, we explore in-hand regrasping for precision grasps (i.e. only contacts with the fingertips). We primarily explore an optimization approach to ``finger gaiting''. In finger gaiting the robot plans motions to change contact points between the object and the fingertips by moving a single finger at a time, while the remaining fingers hold the object stabily. We formulate an optimization problem to find a collision free trajectory for a finger, while the remaining fingers remain fixed. By performing this in order for all fingers we can move the fingers to the desired contact locations on the object, if they lie within the reachable workspace of the current finger pose. However, for many grasps, the robot cannot directly relocate the fingertips to the desired contact locations.

Inspired by the finger gaiting work of Rus~\cite{rus-icra1992} and Leveroni and Salisbury~\cite{leveroni1996reorienting} we add a second component to our in-hand regrasping problem where the robot moves the object relative to its palm, while maintaining the current contact points, in order to move the reachable workspace of its fingers closer to the contact locations defined by the desired grasp. We call this  sub-problem ``in-grasp manipulation''. We build on our previous work for in-grasp trajectory optimization~\cite{sundaralingam2017relaxed} and formulate a separate optimization problem to independently solve this second task.

We then perform regrasp planning by alternatively iterating between solving fingertip relocation for all fingers on the hand and in-grasp object reposing, until the robot achieves the desired grasp. Thus by moving the object, while maintaining the current contact locations, the robot can change the reachable object surface for a single finger. Once at the new pose, the robot can then move this finger closer to the contact point defined by the desired grasp, while the remaining fingers stably hold the object. Fig.~\ref{fig:intro_plan} illustrates an overview of this approach.

To better place our contributions in the broader context of in-hand regrasping, we list some important open problems:
\begin{enumerate}
\item moving to a desired object pose after reaching the desired grasp contact points
\item moving to a set of desired grasp contact points from the current grasp
\item avoiding unwanted collisions between the object and the hand during manipulation
\item ensuring stability of the object  during manipulation
\item choosing the correct sequence of fingers in performing finger gaiting
\item planning an initial grasp, which can achieve the desired object pose through in-grasp re-planning.
\end{enumerate}
In this paper we primarily focus on Problems~2 and 3, moving to a set of desired grasp contact points from an initial grasp while avoiding unwanted collisions. We partially explore grasp stability~(Problem~4), but we do not include dynamics in our approach, so it would be insufficient for execution on a physical robot. Our previous work~\cite{sundaralingam2017relaxed} addresses Problem~1; however, we now present minor extensions for use with finger gaiting. We do not directly address Problems~5 or 6.

As such, the key contributions of this paper are
\begin{enumerate}
\item an optimization based framework for planning finger gaits on arbitrary object meshes, which directly solves for collision-free joint angle trajectories
\item an extension of our previous optimization based framework for in-grasp manipulation~\cite{sundaralingam2017relaxed} to move an object to increase the reachable workspace of a finger, while avoiding unwanted contacts
\item a framework for moving from an initial grasp on an object to a desired grasp using the proposed finger gaiting and in-grasp manipulation optimization methods.
\end{enumerate}

We organize the reminder of the paper as follows. We discuss related work in the next section followed by a formal definition of our problem and proposed approach in Sec.~\ref{sec:prob_def}. We introduce our planner in Sec.~\ref{sec:regrasp-planner}. Implementation details and experimental setup is in Sec.~\ref{sec:exp}. Plans from our approach are discussed in Sec.~\ref{sec:results} followed by concluding remarks in Sec.~\ref{sec:conclusion}.

\section{Related Work}
\label{sec:related_work}
Object regrasping has been mostly explored with respect to using the environment to regrasp using a gripper~\cite{tournassoud1987regrasping} and also by using object dynamics to regrasp the object~\cite{cole1989dynamic,furukawa2006dynamic,cole1992dynamic,dafle-icra2014}. We focus on using the dexterity in the fingers to regrasp and restrict our literature to in-hand regrasping methods.

Literature on in-hand manipulation planning is extensive~\cite{leveroni1996reorienting,omata1996regrasps,goodwine2002motion,goodwine1999stratified,harmati2001object,harmati1999improvedcontrol, harmati2002fitted,cherif1999,cherif2001global,rus1997coordinated,rus1999hand,han1998dextrous,cherif1997planning,mordatch2012a,sundaralingam2017relaxed}. These can be split into two categories in terms of contacts, one where the fingertips always remain in contact~\cite{rus1997coordinated,rus1999hand,cherif1999,cherif2001global,sundaralingam2017relaxed} and methods where fingers break and make new contacts~\cite{leveroni1996reorienting,omata1996regrasps,mordatch2012a}.

Cherif and Gupta define the ``re-configuration'' problem, given the initial grasp and a  desired object pose, as finding a continuous path in the configuration space to a grasp that would reach the desired pose~\cite{cherif1999,cherif2001global}. They propose moving one finger while keeping the other fingers static. They formulate two planners: a high level planner on the configuration space of the object which generates intermediate sub-goals connecting the initial orientation to the desired orientation with no task constraints. The second level planner is a local planner which searches for feasible trajectories to reach the sub-goals. They explore rolling and sliding motions in~\cite{cherif1997planning}. Their method is however limited to smooth convex polyhedra and does not account for breaking of contact. We focus on re-grasping the object by breaking contact and making new contact on the object. We also do not require a smooth polyhedra and only require an object mesh.

Rus's work on coordinated motion planning~\cite{rus1997coordinated,rus1999hand} focuses on planning for object motion with frictionless contacts. \cite{rus1997coordinated} introduces coordinated manipulation of object in two dimensions and focuses on task planning. This is expanded to 3D in~\cite{rus1999hand}. The fingers are split into two sets: fixed and active. Fingers in the active set move using their proposed finger-tracking control while fingers in the fixed set maintain the grasp. They however limit their reconfiguration to a plane. We plan in the full Cartesian space with fingertip contact points on the 3D object mesh.

Leveroni and Salisbury~\cite{leveroni1996reorienting} reorient objects by ``grasp gaits''. They introduce a planner for switching from a current grasp by finger gaiting. They propose using grasp maps for an object, where stable grasp contact regions are marked for two finger grasps and also generate finger workspace maps. Using the generated maps, they setup rules to perform grasp gaits to reorient the object. They formulate the method only for 2-dimensional objects with frictional point contacts. Their method also requires a unique contact point per angle on the object, restricting the object to be convex. 

Omata and Farooqi~\cite{omata1996regrasps} perform regrasping on prism shaped objects with predefined primitives. Given a prism shaped object, they enumerate all possible motions possible with their primitives along the different axes of the object and a search tree is built and used to plan. Their work is limited by the restriction of the primitives to work on specific axes on the object and is not arbitrary. Han and Trinkle~\cite{han1998dextrous}  perform finger-gaiting on a spherical object. They formulate finger gaiting to perform reorientation. However they do not generalize to non-spherical objects and their planner assumes the fingers have a 6D workspace. We optimize in the joint space, ensuring kinematic feasibility for any number of joints.

Finger gait planning has been studied from a stratified motion planning perspective by two groups, Goodwine et. al~\cite{goodwine1999stratified,goodwine2002motion} and Harmati et. al~\cite{harmati1999improvedcontrol,harmati2002fitted}. These methods only focus on moving the object to a desired pose and obtaining fingertip relocations to achieve the task. They do not focus on moving to a goal grasp. They do not explicitly check for collisions between the fingers. In this paper, we focus on moving to a desired grasp which includes moving to goal contact points on the object with constraints to ensure a collision free plan.
\begin{figure*}[]
  \centering
  \begin{tabular}{cccc} \includegraphics[height=4cm]{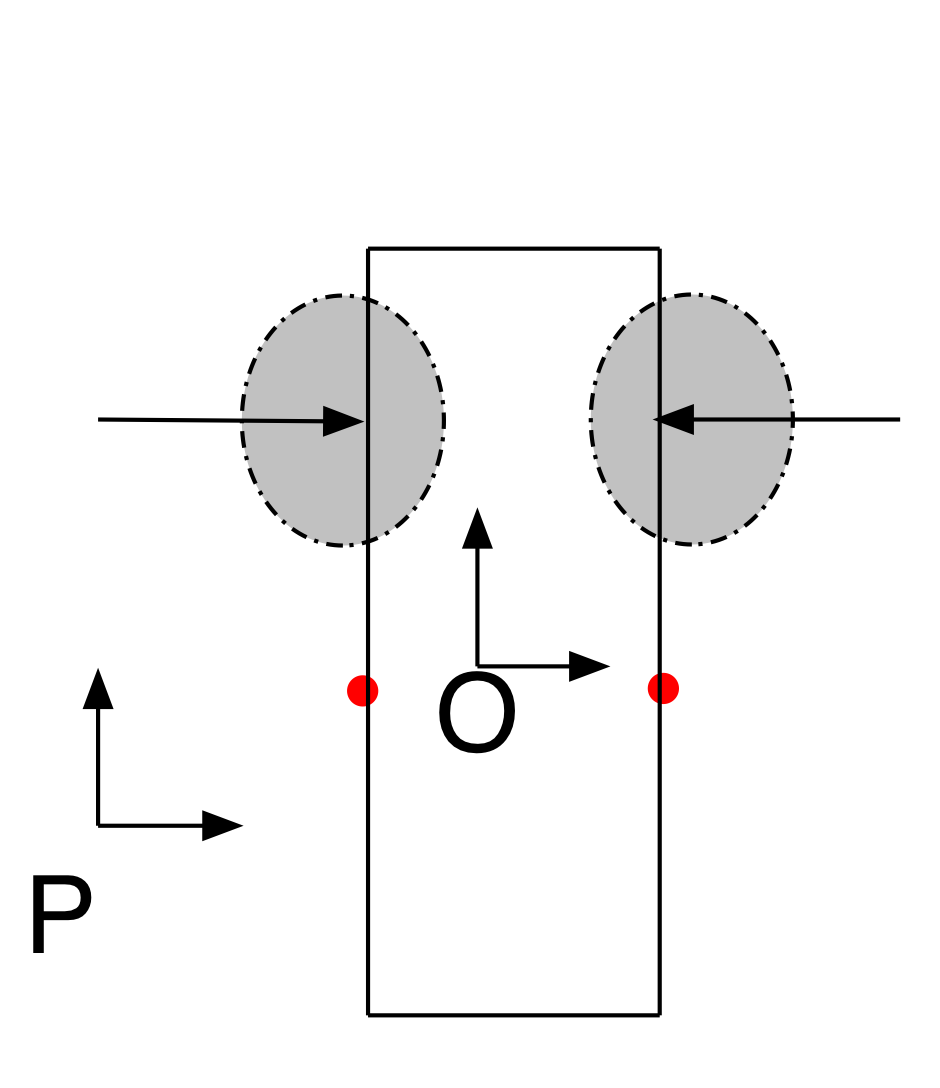}&                                                                \includegraphics[height=4cm]{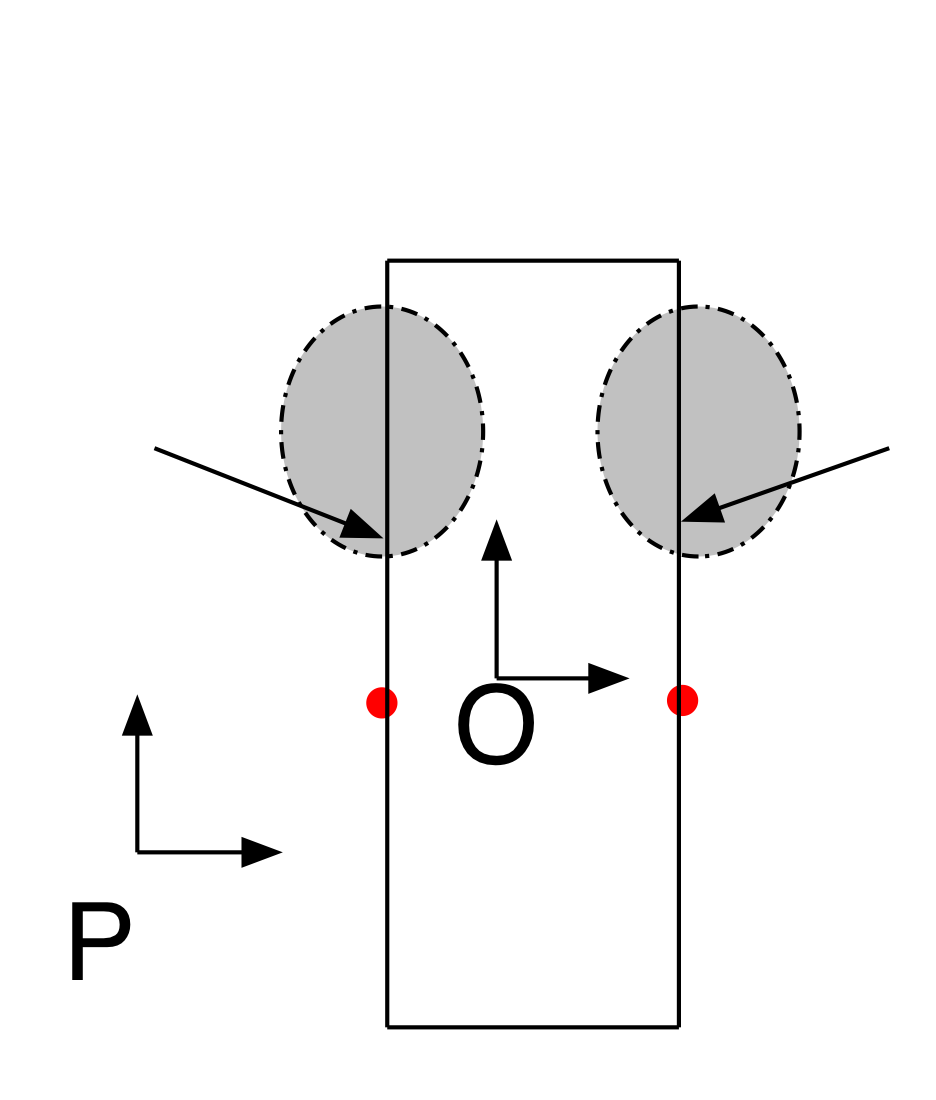}&                                                   \includegraphics[height=4cm]{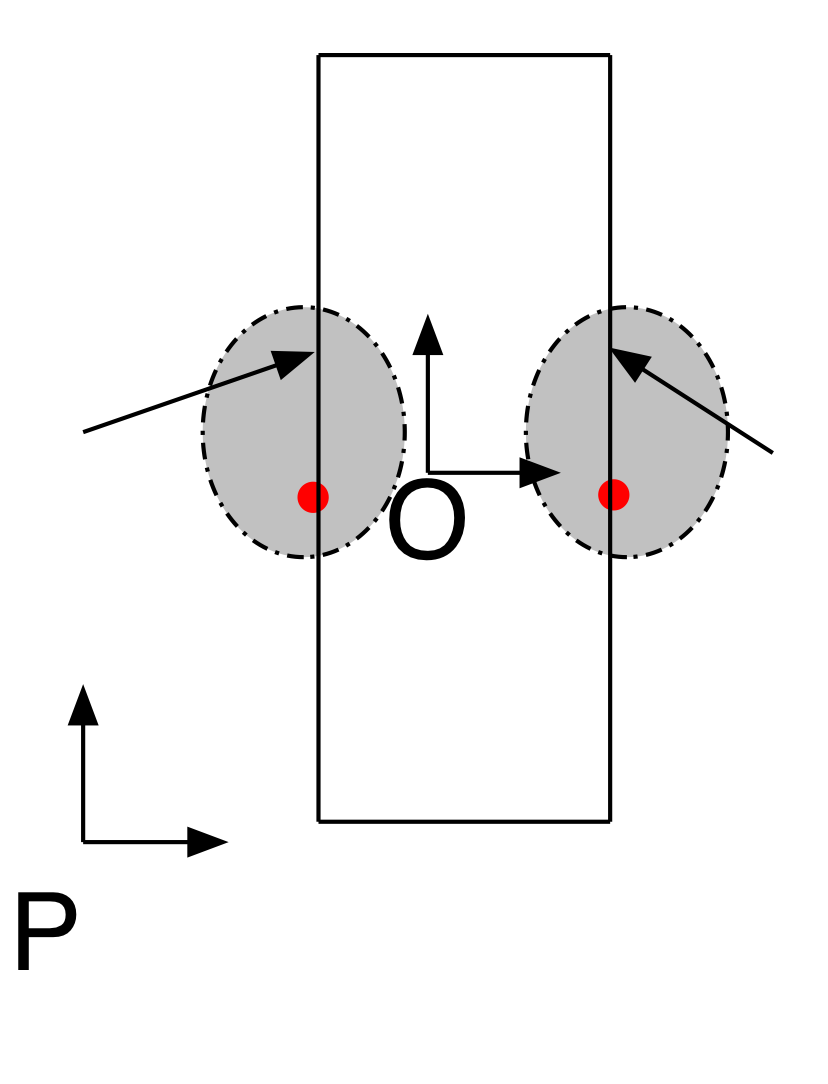}&                                                               \includegraphics[height=4cm]{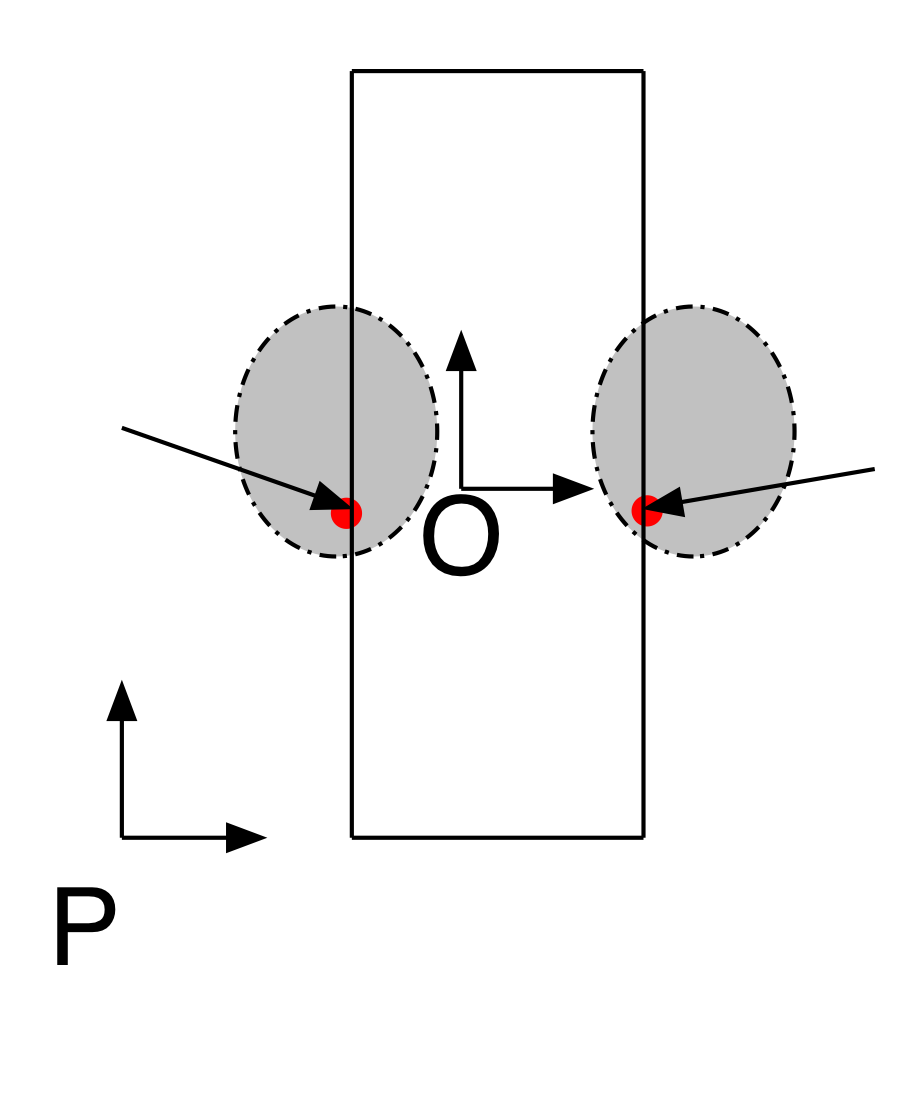}\\      
(a) Initial Grasp & (b) Finger-gaiting & (c) In-grasp manipulation  & (d) Final grasp
    \end{tabular}
  \caption{Steps in our approach to in-hand regrasping are shown in 2D with only two fingers for clarity. The object is in an initial grasp in (a) with reachable workspace of the fingertips shown as gray-shaded ellipses and the contact points for the desired grasp shown as red dots on the object. The palm frame is shown as 'P' and the object frame as 'O'. Finger gaiting is planned within the finger's reachable workspace using OPT1 for the two fingers and they are relocated in~(b), followed by moving the object through in-grasp manipulation~(c) using OPT2. The two steps are iterated until the final grasp is reached, which is shown in~(d). }
  \label{fig:approach}
\end{figure*}

Finger gaiting has been used for grasp stabilization, where a single finger gait is performed to move to a more stable grasp. Buss and Schlegl~\cite{buss1997multi} explore optimizing grasp force to transition from a  $n$ fingered grasp to a $n-1$ finger grasp, allowing the extra finger to break contact with the object. They do not focus on planning a sequence of finger gaits to move to a different grasp. Hang et. al~\cite{hang2016} have shown the effectiveness of finger gaiting for grasp stabilization. They do not perform multiple finger gaits and instead perform single finger gaits as to adapt grasps to maintain the object in the grasp. 

From the literature, it is clear that very little work addresses in-hand regrasp planning for arbitrary object in Full 6-dimensional Cartesian space. We attempt to formulate a generic planner that would allow for in-hand regrasping of objects. We incorporate collision checking as constraints, motivated by recent work in trajectory planning~\cite{ratliff2009chomp,schulman2014}.


\section{Problem definition \& Proposed Approach}
\label{sec:prob_def}
Formally, the problem of in-hand regrasp planning can be defined as finding a sequence of grasps $\mathbf{G}=[G_0,\ldots,G_N]$ which moves the object from an initial grasp $G_0$ to a desired grasp $G_g$ at the final step $N$. Each grasp \(G_i = (X_i, o_i)\) consists of a list of fingertip contact points $X$ and an object pose $o\in SE(3)$. Each contact point list contains $X=[f_1,..,f_m]$, where $f_j\in \mathbb{R}^3$ is the contact point of finger $j$, where $m$ defines the number of fingers on the hand. We also require knowledge of the initial joint configuration of the hand~$\Theta_0$. We approach the problem with the following assumptions:
\begin{enumerate}
\item The object is rigid.
\item The desired grasp is a stable grasp and the desired object pose is reachable at the desired contact points.
\item All grasps considered are precision grasps (i.e. contacts are only made at the fingertips)
\item The order in which the fingers are to be relocated~(gait pattern) is given.
\item We assume that all the fingers can repose their contact on the object. In the case of the thumb, we assume that it can slide to the new contact point.
\end{enumerate}
We split the problem of in-hand regrasp planning into two sub-problems:
\begin{enumerate}
\item Finding a new location for a fingertip within its reachable workspace.
\item Moving the object to shift the reachable workspace of the fingertips relative to the object surface.
\end{enumerate}
We discuss our approach to these two steps in the reminder of this section. Sec.~\ref{sec:regrasp-planner} combines these two sub-components into a single in-hand regrasp planner. An overview of our approach is illustrated in Fig.~\ref{fig:approach}. For convenience, we summarize the symbols we use in Tab.~\ref{tab:notations}.
\subsection{Optimization for Finger Gaits~(OPT1)}
\label{sec:find-new-fing}
This sub-problem finds the contact point~$f_{r,j}$ at step $j$ for finger $r$, in the reachable workspace~$R_r$ of finger $r$ that moves the fingertip towards the goal finger contact point $f_{r,g}\in X_g$. We formulate this step as a constrained geometric optimization problem over the joint angles \(\Theta^r\) of finger \(r\), while the remaining joints in \(\Theta\) remain fixed.  The cost function, Eq~\ref{eq:opt1}, penalizes the distance between the desired contact point $f_{r,g}$ and the fingertip location planned as a function of the hand's joint angles.
\begin{flalign*}
  \min_{\Theta^r}\hspace{5pt}  &  D(f_{r,g},FK_r(\Theta^r))\numberthis\label{eq:opt1}\\
  \text{s.t.}\hspace{5pt}    & \\
  &\Theta_{min}^r \preceq \Theta^r \preceq\Theta_{max}^r \numberthis\label{eq:position_limits}\\
  & SD(FK_r(\Theta^r),M)=0\numberthis\label{eq:surface_const}\\
  & C(\Theta^r,M)=0 \numberthis\label{eq:collision}\\
  & S(FK_r(\Theta^r)) \leq \eta, \numberthis\label{eq:stability_current}
\end{flalign*}
\begin{table}
  \centering
  \caption{Symbols}
  \begin{tabular}{cl}\toprule
    \textbf{Symbol}   & \textbf{Description}\\ \hline
    $m$        & Number of fingers\\
    $N$        & Number of steps \\
    $M$        & Object mesh\\
    $F$        & list of fingers \\\midrule
    $\Theta_j$ & Robot hand joint configuration at step $j$\\
    $o_j$      & Object pose at step $j$\\
    $X_j$      & List of contact points at step $j$\\ \midrule
    $P$        & Ordered list of finger gait pattern \\
    $R_i$      & Reachable Workspace of finger-i\\
    $f_{r,j}$    & Contact point of finger-$r$ at step $j$\\
    $f_{r,g}$    & Goal contact point of finger $r$  \\
    $L_r$ & Links in a finger $r$\\\midrule
    \bottomrule
  \end{tabular}
  \label{tab:notations}
\end{table}
The function~$FK_r(\cdot)$ computes the pose of fingertip of finger~$r$. The joint limit constraints defined in Eq.~\ref{eq:position_limits} ensure kinematic reachability for the fingertip.
The constraint in Eq.~\ref{eq:surface_const} computes the signed distance \(SD(\cdot)\) to ensure the contact point of the fingertip lies on the surface of the object mesh $M$. The signed distance computes the shortest distance between a point \(p\) and the mesh \(M\). The sign denotes if \(p\) lies within the mesh (negative) or outside the mesh (positive). The constraint in Eq.~\ref{eq:collision} ensures the finger links do not collide with the object at any point other than the fingertip.

We compute the collision cost (Eq.~\ref{eq:collision}) as:
\begin{align*}
  C(\Theta^r,M)&=\sum_{l\in L_r}(\beta-\min(\beta,SD(FK_l(\Theta^r),M))) \numberthis
\end{align*}
which ensures all links (excluding the fingertip) on the moving finger maintain at least distance of \(\beta\) from the object. The function~$FK_l(\cdot)$ computes the pose of link~$l$.

The constraint in Eq.~\ref{eq:stability_current} ensures the grasp remains stable during finger gaiting and at the resulting grasp~$G_{j+1}$.
While any grasp stability measure could be used in theory, we formulate a simple measure to approximate this stability. We simply limit the finger gait distance to be within a threshold, implying that the resulting grasps will be similar to the current grasp, which initially is known to be stable.
\begin{align*}
  S(FK(\Theta^r))&=||FK_r(\Theta^r_0)-FK_r(\Theta^r)||_2^2\numberthis\label{eq:stability}
\end{align*}
where $\Theta_0^r$ define the joint angles of the finger prior to the optimization. This ensures only small steps are taken when $\eta$ is small.

We define our cost function as a generic distance, but focus on evaluating the Euclidean distance in this work:
\begin{align*}
D(f_{r,g},FK_r(\Theta^r))&= ||f_{r,g}-FK_r(\Theta^r)||_2^2 \numberthis
\end{align*}
\subsection{Optimization for Object Reposing~(OPT2)}
\label{sec:moving-object}
This sub-problem focuses on moving the object given fixed contact locations, in order to shift the reachable workspace of the fingertips relative to the object.
We approach this problem using in-grasp manipulation. Our previous work~\cite{sundaralingam2017relaxed} showed how to move to a desired object pose within the currently reachable workspace of the fingers. We modify this slightly for use here, by changing the first component of our cost function \(E_{des}\) to move to improve the reachable workspace for finger gaiting instead of moving towards a single desired pose.
We additionally add collision constraints (Eq.\ref{eq:palm_collision}) and simplify the problem to optimize only for a final joint configuration instead of a full joint trajectory.

We formulate the full optimization as:
\begin{flalign*}
\min_{\mathbf{\Theta}}\hspace{5pt} & E_{des} +k_1E_{pos}(\Theta)+k_2E_{or}(\Theta)\numberthis\label{eq:costs}\\
 \text{s.t.}&\\
 &\Theta_{min}\preceq \Theta \preceq\Theta_{max} \numberthis\label{eq:position_limit}\\
 & C(\Theta,M)=0 \numberthis\label{eq:palm_collision}
\end{flalign*}

The first constraint enforces the joint position limits of the robot hand and the second constraint enforces the object to not collide with the hand. The cost terms \(E_{pos}\)and \(E_{or}\) define the relaxed-rigidity constraint, encouraging fingertips to keep the same contact locations on the object as in the initial grasp, while allowing for slight sliding and rolling at these contacts.
The scalar weights, $k_1,k_2$, on each cost term allow us to trade-off between the three cost components. For more details of the planner see~\cite{sundaralingam2017relaxed}.
The cost term $E_{des}$ moves the object so that the finger gaiting optimization can place fingers closer to the desired grasp. There are multiple ways to formulate this cost term and we explore two formulations.

The first formulation reduces the distance between the reachable workspace of the fingers \(R_{r\in[0,m]}\) and the desired contact points \(f_{r\in[0,m],g}\). We define this cost as a sum over costs for each finger. For a given finger \(r\) we penalize the desired contact location \(f_{r,g}\) lying outside of the fingertip's reachable workspace \(R_r\) (represented as a convex mesh). We compute this as the maximum between 0 and the signed distance between the desired contact point and the boundary of the reachable workspace mesh:
\begin{align*}
E_{des}&=\sum_{r=0}^{m}\max(0,SD(f_{r,g},R_r)) \numberthis
\end{align*}

In our second formulation we first solve an auxiliary optimization, finding the object pose, \(\hat{O}_d\), which minimizes the Euclidean distance between the current grasp contact points and the desired grasp contact points. We compute this minimizing transform using singular value decomposition as explained in~\cite{besl1992method}.
We can then set $\hat{O}_d$ as the desired object pose and directly minimize the object's pose error using the cost function from our previous work~\cite{sundaralingam2017relaxed}:
\begin{align*}
E_{des}&=E_{obj}(\Theta,\hat{O}_{d}) \numberthis
\end{align*}
\section{Regrasp Planner}
\label{sec:regrasp-planner}
We now formulate a planner for the in-hand regrasping problem, leveraging the two optimization problems, presented in the previous section. As a reminder, we define the goal of in-hand regrasping as moving to a desired grasp \(G_d = (X_d, o_d)\) without dropping the object.
We assume a fixed gait pattern $P$, since searching over the gait pattern is in itself a complex problem. We present the algorithm pseudocode in Alg.~\ref{alg:regrasp}.
\IncMargin{1em}
\begin{algorithm}
\SetAlgoLined
\KwData{$M$,$K_0$,$X_g$,$o_g$}
\KwResult{$\mathbf{K}=[G,\mathbf{\Theta}]$}
\textbf{K}$\,=\,$[]\;
\textbf{K}.append($K_0$)\;
$\text{err} \leftarrow \max_{r\in[0,m]}(f_{r,0}-f_{r,g})$\;
$n=0$\;
\While{$\text{\upshape err}>\zeta$ and $n<50$}{
  \For{i$\in$P}{
    $K_t \leftarrow$ OPT1(\textbf{K}.last, $X_g$, $i$)\;
    \textbf{K}.append($K_t$)\;
  }
  $\text{err} \leftarrow  \max_{r\in[0,m]}(f_{r,t}-f_{r,g})$\;
  \If{$\text{\upshape err}>\zeta$}
  {
    $K_t$ $\leftarrow$ OPT2(\textbf{K}.last,$X_g$)\;
    \textbf{K}.append($K_t$)\;
  }
  $n{+}{+}$\;
  }
  $K_t \leftarrow \;$in\_grasp(\textbf{K}.last, $o_g$)\;
  \textbf{K}.append($K_t$)\;
  return \textbf{K}\;
\caption{In-hand Regrasping Planner\label{alg:regrasp}}
\end{algorithm}
\DecMargin{1em}

Given the initial grasp, the desired contact points, and the gait pattern \(P\), we first plan finger gaiting using OPT1 following the finger order of $P$~(lines 5-8) and add the new grasp to the grasp sequence \textbf{K}. OPT2 then reposes the object. These steps iteratively alternate until the error is less than~$\zeta$. The grasp sequence \textbf{K} also stores the joint configurations $\Theta$ at every sequence step.
Once the plan reaches the desired contact locations \(X_d\), we perform a final optimization using our in-grasp manipulation planner from~\cite{sundaralingam2017relaxed} to move the object to the desired pose \(o_d\) (lines 16-17).


\section{Experimental setup and Implementation details}
\label{sec:exp}
\begin{figure}
  \centering
  \includegraphics[width=0.45\textwidth]{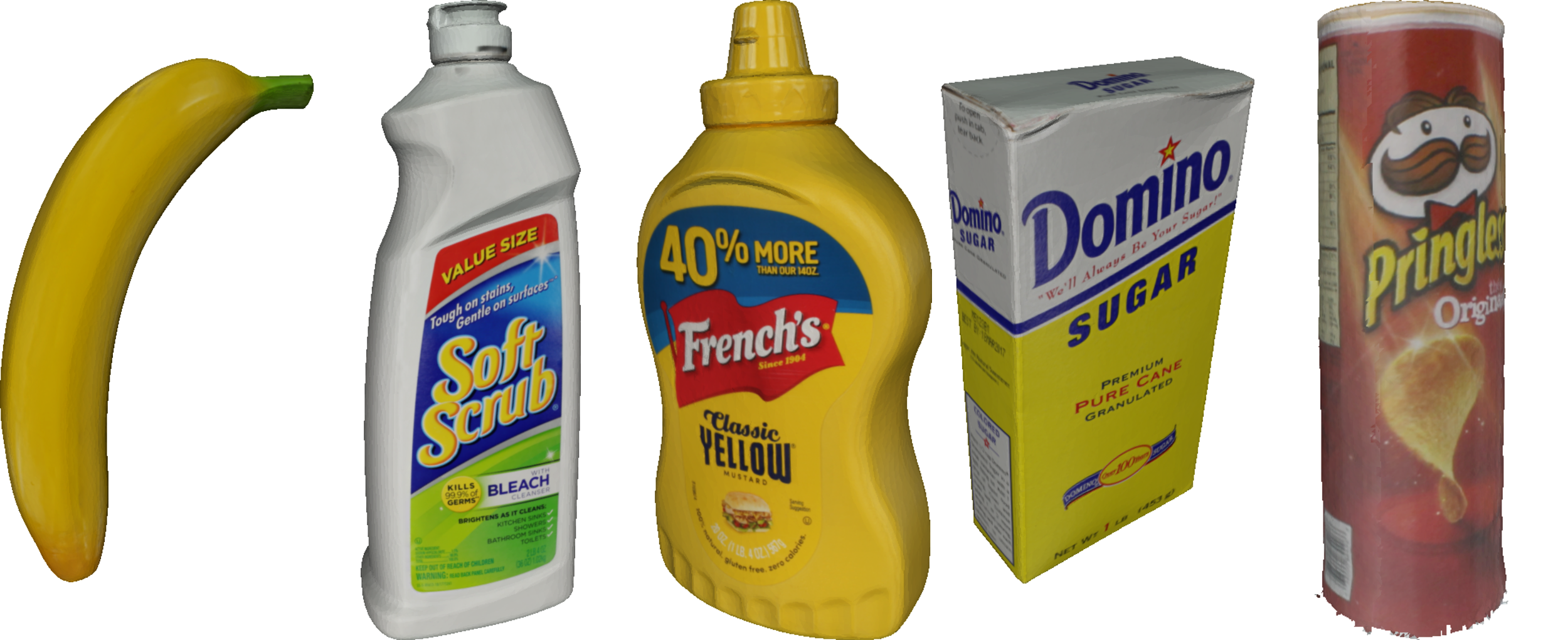}
  \begin{tabular}{ccccc}
    Banana & Soft-scrub & Mustard &Sugar-box & Pringles
  \end{tabular}
  \caption{Objects tested with our planner.}
  \label{fig:objects}
\end{figure}

The optimization frameworks are implemented as sequential quadratic programs~(SQP's) with analytic gradients for the cost terms and the constraints. We use SNOPT~\cite{gill2005}, an SQP solver to perform the optimization in the Pagmo framework~\cite{biscani2010global}. We perform experiments using the Allegro hand\footnote{http://www.simlab.co.kr/Allegro-Hand.htm} in simulation to evaluate our in-hand regrasping planner. Objects are chosen from the YCB dataset~\cite{calli2015}. We show the chosen objects with their labels in Fig.\ref{fig:objects}. 
Computations are performed on a computer with an Intel i7-7700k processor with 32 GB of RAM running Ubuntu 16.04.  We compute signed distances using libccd\footnote{https://github.com/danfis/libccd} based on a combination of the Gilbert-Johnson-Keerthi~(GJK) algorithm and the expanding polytope algorithm~(EPA), extensive details are found in~\cite{van2001proximity}. We approximately decompose non-convex objects into convex groups using~\cite{mamou2009simple} to speedup signed distance computation. We compute the reachable workspace of the fingertips using voxel-based workspace estimation~\cite{anderson2012voxel}. With a mesh for the reachable workspace, we can compute the signed distance between the desired contact point and this mesh. We obtained laserscan meshes for ``Banana'', ``Mustard'', ''Soft-scrub'' and ``Sugar-box''; for the ``Pringles'', we used a lower accuracy mesh obtained from an RGB-D sensor. All associted software and data is available at~\url{https://robot-learning.cs.utah.edu/project/in_hand_manipulation}.

We generate initial and desired grasps manually. All generated grasps are four-fingered precision grasps. We generate 5 pairs of initial and desired grasps per object to evaluate our planner.  We use two gait patterns based on the desired contact location of the index fingertip. If the desired contact location is farther from the middle finger than the index finger, we use the gait pattern \{index, middle, ring, thumb\}. In the case that the desired contact location is closer to the middle finger, we use the gait pattern \{thumb,ring,middle,index\}. We set this gait pattern before we start the planner and do not change during planning.

For the generated in-hand regrasping plans, we report, the average error between the desired grasp contact points and the planned final grasp contact points, the computation time for generating the plans and the number of iterations our planner runs until convergence~(error less than $\zeta$).  We limit the maximum number of iterations to 50. We compute the error between the planned and desired final grasp contact grasp contact points using Euclidean distance between the contact point pairs and average over all the fingers. We report this error as ``Average Point Error'' in the following sections. Our approach to OPT2 has two formulations, we term the first formulation which reduces the signed distance between the reachable workspace and the desired contact points as ``SD''. The second formulation which uses SVD to find a rigid transformation for the object pose is termed as ``SVD''. We report results for both of these formulations. The values of~$\eta$ and~$\beta$ in OPT1 are chosen as $1cm$ and $0.1cm$ respectively. The weights $k_1$ and $k_2$ in OPT2 are chosen to be 1000 and 10 respectively. The threshold~$\zeta$ is chosen to be $6mm$.

\section{Results}
\label{sec:results}
\begin{figure*}
  \centering
  \begin{tabular}{c c c}
    \multirow{ 2}{*}{\rotatebox{90}{Mustard}} & \rotatebox{90}{\hspace{2.5em} SD}  & \includegraphics[width=0.9\textwidth]{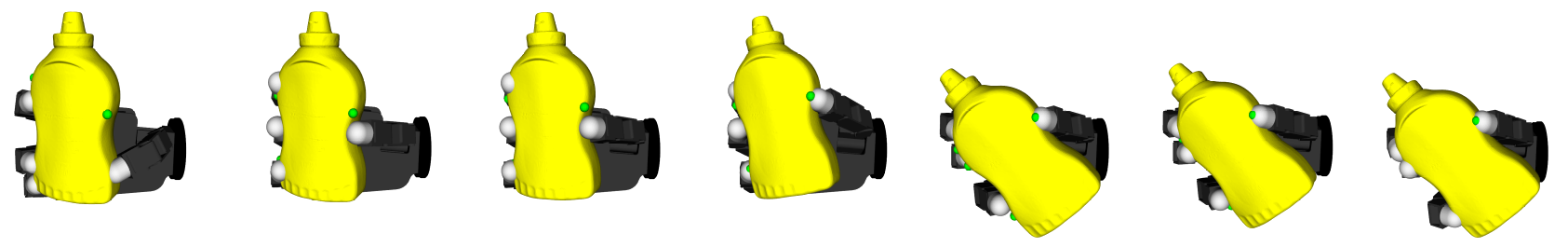}\\
  &\rotatebox{90}{\hspace{1.5em} SVD} & \includegraphics[width=0.9\textwidth]{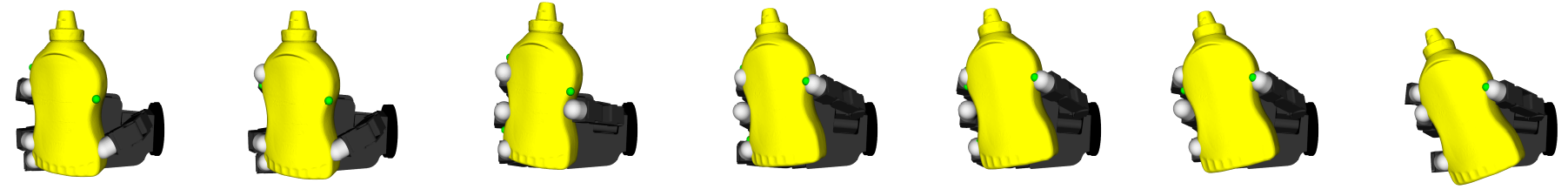}\\ \hline
\multirow{ 2}{*}{\rotatebox{90}{Soft-scrub}}   & \rotatebox{90}{\hspace{2.5em} SD} & \includegraphics[width=0.9\textwidth]{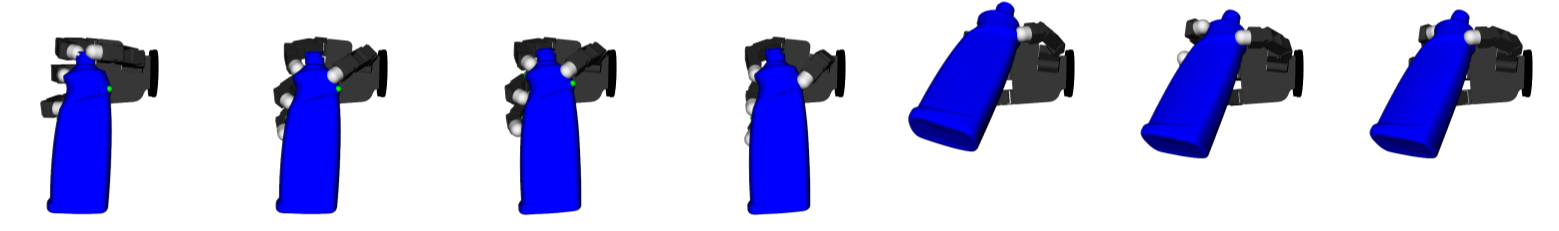}\\ 
  & \rotatebox{90}{\hspace{2.5em} SVD}& \includegraphics[width=0.9\textwidth]{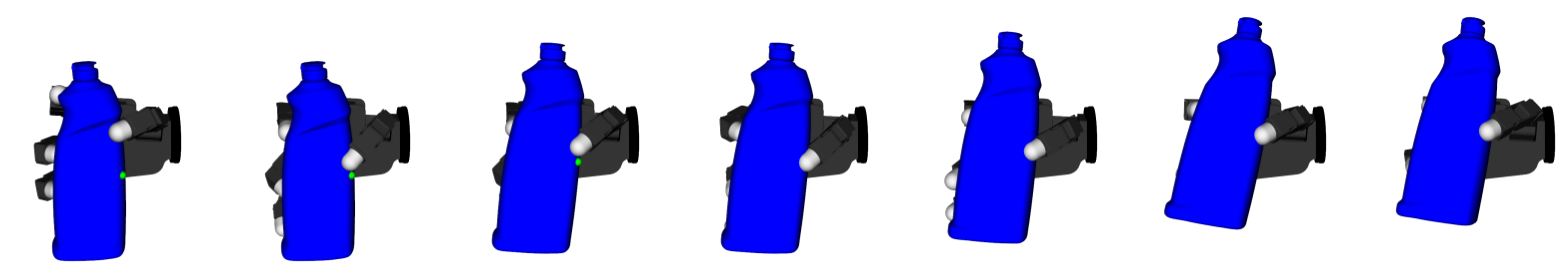}\\ \hline
\multirow{ 2}{*}{\rotatebox{90}{Banana}}   & \rotatebox{90}{ \hspace{2.5em} SD} & \includegraphics[width=0.9\textwidth]{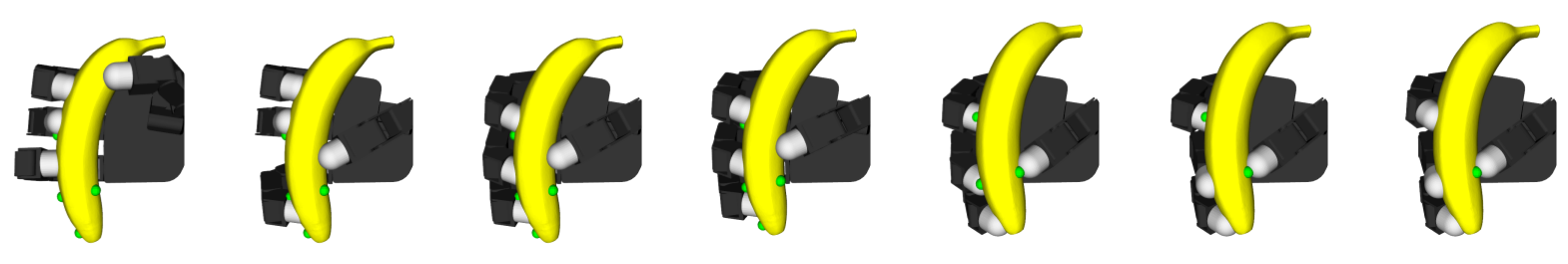}\\ 
  &\rotatebox{90}{\hspace{2.5em} SVD} & \includegraphics[width=0.9\textwidth]{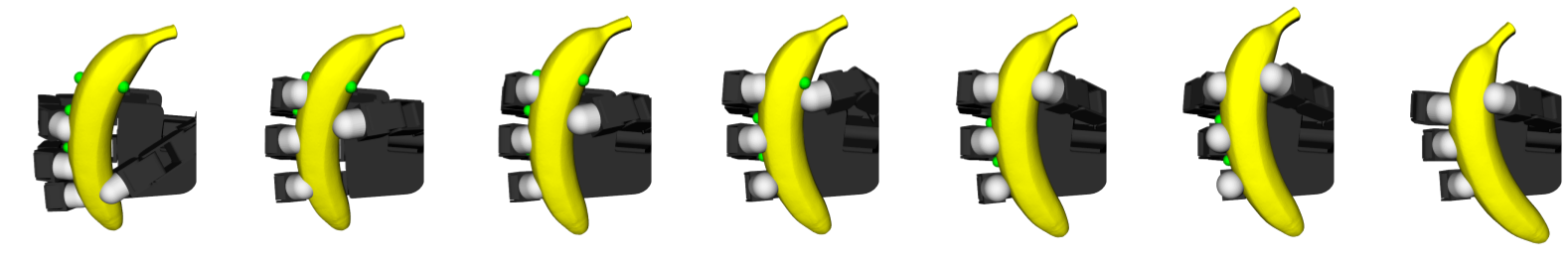}\\
    \end{tabular}
  \caption{Sample plans from our in-hand regrasping approach with green dots showing the desired contact points. The plans with ``soft-scrub'' object show how the in-grasp manipulation lifts the object up to reach for the desired contact points. }
  \label{fig:plans}
\end{figure*}

We first discuss the convergence rate of our planner, followed by planning time and the obtained final grasp errors. Some plans for the objects are shown in Fig.~\ref{fig:plans}. 

\subsection{Alternating optimization feasibility}
\begin{figure}
  \centering
  \includegraphics[width=0.38\textwidth]{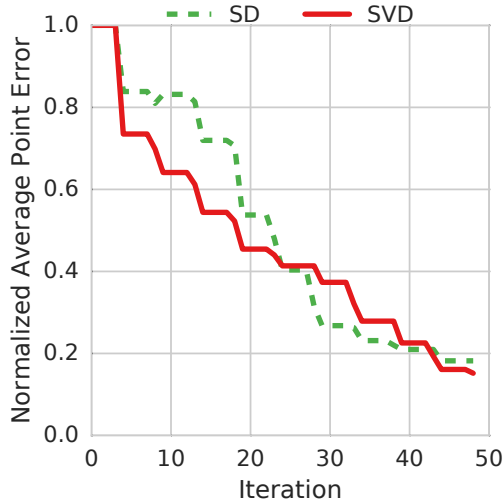}
  \caption{The normalized ``Average Point Error'' across all the object is shown here. Since we limited our iteration to 50 and also have a threshold~$\zeta$ on the final contact point error, the convergence does not reach zero.}
  \label{fig:convergence}
\end{figure}

Our iterative approach to in-hand regrasp planning disconnects the reposing and finger gaiting optimization methods. We approach the planner in a greedy scheme, hence it is essential to study the ``Average Point Error'' after every iteration in the planner. Fig.~\ref{fig:convergence} shows the normalized ``Average Point Error'' over all of the generated plans for ``SD'' and ``SVD'' methods. The error decreases after every iteration, indicating the effectiveness of our planner. ``SVD'' converges faster than ``SD'' initially, as the rigid transformation gives a better initial object pose estimate. After 25 iterations, the  ``SD'' convergence rate increases since at-least one finger has reached the desired contact point, at this time.
\subsection{Planning time}
We report the time taken between initializing the planner at the initial grasp, and when the final grasp plan is obtained as the planning time. The ``SD'' method was computationally intensive as the cost function in OPT2 had to minimize the signed distance for multiple fingers, produces a median planning time over all the objects of 729.05 seconds. However, ``SVD'' was only minimizing the error in the object pose, making the planning time much faster than ``SD'' with a median planning time over all the objects of 75 seconds. Approximately, 10x improvement in planning time was seen with the ``SVD'' method. Planning times per object are reported in Tab.~\ref{tab:results}. While the planning time was drastically decreased with ``SVD'', the number of iterations increased with ``SVD''. ``SVD'' took a median of 44 steps across all objects while ``SD'' took only 29 steps. This reflects the effect of ``SVD'' in speeding up in-hand regrasp planning even when taking more iterations. An interesting comparison is in the plans obtained for the ``mustard'' object, shown in Fig.~\ref{fig:plans}. ``SD'' method translates the object lower and rotates the object to reach the desired contact points, while ``SVD'' lifts the object first and then rotates slightly to reach the desired contact points. The ``Pringles'' object mesh being lower quality, took longer to plan, with one plan taking 3513.96 seconds. 
\begin{table}
  \centering
  \caption{Summary of planning time across all the object.}
\label{tab:results}
\begin{tabular}{|l|c|c|c|}
\hline
\textbf{Object}              & \textbf{Method} & \textbf{ Maximum(s)} & \textbf{Median(s)} \\ \hline
  \multirow{2}{*}{Banana}      & SD              &           1025.35 &  902.65 \\ \cline{2-4} 
                             & SVD             &        84 & 45            \\ \hline
  
  \multirow{2}{*}{Sugar-box}  & SD              &         2283.37 &  1917.41      \\ \cline{2-4} 
                             & SVD             &      112.59 & 99.38                  \\ \hline
  
  \multirow{2}{*}{Mustard}     & SD              &    1133.504 & 649.69         \\ \cline{2-4} 
                             & SVD             &    135 & 92.83                                         \\ \hline
  
\multirow{2}{*}{Soft-scrub} & SD              &        884.927 & 388.317                               \\ \cline{2-4} 
                             & SVD             &            95 & 74.97                   \\ \hline
  
\multirow{2}{*}{Pringles}    & SD              &   3513.96 & 796.93                                 \\ \cline{2-4} 
                             & SVD             &   134.275 & 16.0052                                                     \\ \hline
\end{tabular}
\end{table}
\subsection{Reaching Desired Contact Points}
\begin{figure}
  \centering
  \includegraphics[width=0.42\textwidth]{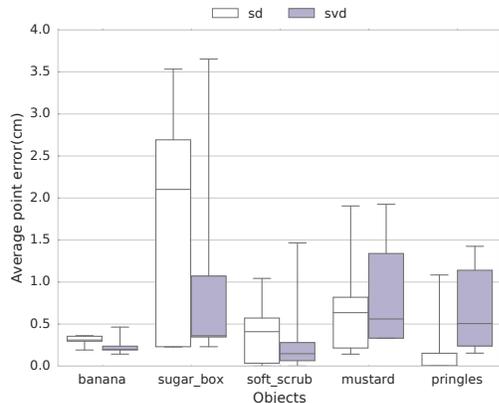}
  \caption{``Average Point Error'' across the five objects is shown. Median error in ``Sugar-box'' improves drastically with ``SVD'' as it can find a transformation more efficiently than the ``SD'' method.}
  \label{fig:errors}
\end{figure}

We now discuss the error in reaching the desired contact points. Fig.~\ref{fig:errors} shows the ``Average Point Error'' across all objects for the two methods. 
We see that the error is lowest for ``Banana'' and largest for the ``sugar-box''. This is partly because the ``Banana'' being smaller, has a large reachable fingertip surface area, making finger gaiting cover larger distances. The median error is lower in all objects except ``Pringles''.  The ``SD'' method lacks any global information about traversing through the edges of a large object and this is reflected in the large error for the ``Suger-box''. ``SVD'' method moves the median error for the ``Sugar-box'' to 0.36 cm as the rigid transformation gives OPT2 a better estimate of the object pose that would move the reachable workspace of the fingertips towards the desired grasp.

\section{Discussion and Future Work}
\label{sec:conclusion}
We presented an optimization based planner that can generate collision-free plans to regrasp objects in-hand. Our decomposed formulation of the in-hand regrasp planner allows for using other in-hand manipulation primitives such as pivoting in addition to in-grasp manipulation and finger gaiting.

However, this paper highlights the challenges present in performing in-hand regrasping of arbitrary objects and the assumptions that have to be proposed to explore the problem. Solving the entire problem of in-hand regrasping remains a long term research problem and as such we attempt to solve parts of the in-hand regrasping problem. Our future work will involve relaxing these assumptions.

While the use of Euclidean distance in our finger gaiting cost function leads to good results in practice, we believe the geodesic distance, which calculates the shortest path distance between the two points on the mesh, would provide some benefits for in-hand regrasping. A motivating example would be moving from a grasp at one end of a 'U' shaped object to the other end. While the initial and final grasp points are close in terms of Euclidean distance, attempting to optimize using this measure results in the planner getting stuck in a local minimum. However, the geodesic correctly shows that the robot must finger gait along the entire distance of the object to reach these points. While this appears to be the more appropriate cost function, efficiently optimizing over the function leads to a much harder problem, requiring non-trivial discrete differential geometry. We are actively working to involve this into our optimization.

Our conservative constraint on the grasp stability causes our regrasp planner to take many small fingertip relocations to reach the final grasp. A significant reduction in the number of required steps by increasing the finger gait distance~$\eta$ was observed. We will explore different values for this parameter as part of validation of our planner on a physical robot.
\section*{Acknowledgment}
B.~Sundaralingam was supported in part by NSF Award \#1657596.
\bibliography{references_planning,references_bala}
\end{document}